\pdfoutput=1

\documentclass[11pt]{article}

\usepackage[]{acl}

\usepackage{times}
\usepackage{latexsym}
\usepackage{tabularx}
\usepackage{colortbl}
\usepackage{arydshln}
\usepackage{xspace}

\usepackage[T1]{fontenc}

\usepackage[utf8]{inputenc}

\newcommand{\approach}{{\textsc{BeInfo}}\xspace}
\usepackage{microtype}

\newcommand{\rparagraph}[1]{\vspace{1.2mm}\noindent\textbf{#1.}}

\newcommand{\sparagraph}[1]{\vspace{0.0mm}\noindent\textbf{#1.}}

\newcommand{\sparagraphnodot}[1]{\vspace{0.0mm}\noindent\textbf{#1}}

\newcommand{\mustardcell}{\cellcolor[HTML]{FFE599}}

\newcommand{\peachcell}{\cellcolor[HTML]{FCE5CD}}

\definecolor{Gray}{gray}{0.95}
\definecolor{racing-green}{rgb}{0.0, 0.8, 0.6}
\definecolor{awesome-red}{rgb}{1.0, 0.13, 0.32}
\newcolumntype{Y}{>{\centering\arraybackslash}X}

\usepackage{todonotes}
\makeatletter
\newcommand*\iftodonotes{\if@todonotes@disabled\expandafter\@secondoftwo\else\expandafter\@firstoftwo\fi}
\makeatother

\usepackage{inconsolata}

\usepackage{booktabs}
\usepackage{multirow}

\title{\textit{Dial \approach for Faithfulness}: Improving Factuality of Information-Seeking Dialogue via Behavioural Fine-Tuning}

\author{
Evgeniia Razumovskaia\thanks{~~LTL, University of Cambridge. Work conducted during the internship at PolyAI.}, Ivan Vuli\'{c}, Pavle Markovi\'{c}, Tomasz Cichy,\\
{\bf Qian Zheng, Tsung-Hsien Wen \textnormal{and} Pawe\l{} Budzianowski}
 \vspace{2mm} \\
 PolyAI Limited \\
 London, United Kingdom \\
 \url{poly.ai}
}

\begin{document}
\maketitle
\begin{abstract}
Factual faithfulness is a crucial requirement in information-seeking dialogue: the system should respond to the user queries so that the responses are meaningful and aligned with the knowledge provided to the system. However, most modern large language models (LLMs) suffer from hallucinations, that is, they generate responses not supported by or even contradicting the knowledge source. To mitigate the issue and increase faithfulness of information-seeking dialogue systems supported by the LLMs, we introduce \approach, a simple yet effective method that applies `behavioural tuning' on the LLMs to aid information-seeking dialogue. Relying on three standard information seeking dialogue datasets, we show that models tuned with \approach become considerably more faithful to the knowledge source both for datasets and domains seen during \approach-tuning, as well as on unseen domains, when applied in a zero-shot manner. In addition, we present a `real-life' case study on conversations with real users, showcasing that the models with 3B parameters (e.g., Flan-T5) tuned with \approach demonstrate strong performance on data from real `production' conversations: when tuned on a limited amount of such realistic in-domain dialogues, they surpass much larger LLMs used `off-the-shelf', both on automatic and human evaluation metrics.
\end{abstract}

\section{Introduction}

Pretrained large language models (LLMs), being able to generate natural and grammatical text and respond coherently to user queries, are the mainstay of modern NLP~\citep{naveed2023comprehensive}. They have demonstrated their capabilities in a plethora of tasks where the general world knowledge, which can be learnt via pretraining directly from the data, is required \citep{touvron2023llama, hoffmann2022chinchila}. However, reliance only on the content  from the pretraining data also means that the model's responses might be generic or not be up to date, especially for queries responses to which change across time such as \textit{Who is the current prime minister of the United Kingdom?} An even more prominent issue is \textit{hallucination}~\citep{zhang2023-sirens-in-hallucinations}, a phenomenon often observed even with the most powerful LLMs: the models are prone to output incoherent, irrelevant and/or even factually incorrect or unsupported statements~\citep{naveed2023comprehensive}.


A widely used method to ground and control the content of the output of an LLM is \textit{retrieval-augmented generation} (RAG; \citeauthor{lewis2020RAG}, \citeyear{lewis2020RAG}), where the input to the model is complemented with a retrieved external knowledge source relevant to the user's query. However, even with the use of RAG, the model's output can be unpredictable and not fully controllable: they still sometimes do not adhere to the knowledge source and hallucinate \cite{shuster-etal-2021-retrieval-augmentation}, which can decrease their applicability in user-facing scenarios, as well as raise concerns of their safety~\cite{daheim2023elasticwr}.

\begin{figure}[!t]
    \centering
    \includegraphics[width=0.48\textwidth]{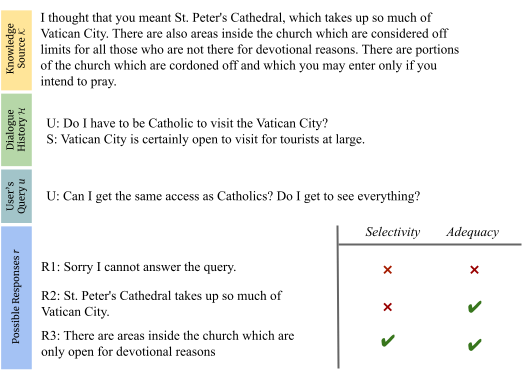}
    \caption{An example of an information-seeking dialogue based on the DoQA dataset \cite{campos-etal-2020-doqa}. Potential responses R1, R2, R3 at the bottom illustrate different issues with two crucial aspects of factual faithfulness: selectivity and response adequacy.}
    \label{fig: dialogue example}
    \vspace{-1.5mm}
\end{figure}


The problem of adherence to the knowledge sources is especially important in the context of \textit{information-seeking dialogue}~\cite{saeidi-etal-2018-interpretation}. The core of this task is to maintain a conversation with the user and respond to their queries based on the provided knowledge source. Figure \ref{fig: dialogue example} presents an example of information-seeking dialogue between the user and the system and potential responses of the system. Orthogonally to improving retrieval systems themselves~\cite{wang2023learningretrieverforrag,kdd2023}, prior work has attempted to combat hallucinations with task arithmetic~\citep{daheim2023elasticwr}, conditioning generation on special control tokens \cite{rashkin-etal-2021-increasing}, and by incorporating a token-level critic which judges the faithfulness of the generated response \cite{dziri-etal-2021-neural}. However, the proposed approaches requires either training an additional model or using complex inference processes such as context-aware decoding \cite{shi2023trusting}.

In this work, we propose \textbf{\approach}, a simple yet effective method that applies `\textbf{be}havioural fine-tuning' of LLMs to increase faithfulness of the generated responses for \textbf{info}rmation-seeking dialogue supported by the LLMs. The model is tuned on a reasonably sized collection of publicly available dialogue data with the true knowledge source(s) extended with randomly sampled facts from a large knowledge base. Intuitively, this should teach the model to become more selective in the information it uses to generate the response and `prepare' its expected behaviour (hence the term `behavioural tuning') for the intended task of knowledge-grounded dialogue. The tuned model can either be used `as is' or as a starting point to fine-tune it further to a specific domain. 


First, we assess the effectiveness of \approach on three standard datasets for information-seeking dialogue:
FaithDial, TopiOCQA and DoQA. Our results demonstrate that \approach leads to consistent improvements in factual faithfulness across several standard evaluation metrics, also with on par or larger lexical overlap between the generated and golden responses. The improvements are especially pronounced when models tuned with \approach are applied in a zero-shot manner to unseen datasets and domains, indicating the usefulness of behavioural tuning for the task. We then present a case study focused on conversations with real users: the main result demonstrates that combining \approach with a small number of in-domain dialogues can substantially increase dialogue factuality even in specialized dialogue domains. The code for \approach is available online at: \url{[URL-ANONYMOUS]}.


\section{Methodology}

 \sparagraph{Task Definition} The aim of information-seeking dialogue is to provide the user with information they need based on one or more knowledge sources, which are typically retrieved from a large knowledge base. More formally, given the knowledge source $\mathcal{K}$, the dialogue history $\mathcal{H}$ and the user's query $u$, the system should output the response $r$ which is factually faithful to $\mathcal{K}$. Here, we follow \citet{rashkin-etal-2021-increasing} and \citet{dziri-etal-2022-faithdial}'s direct definition of \textit{faithfulness}: the response should not contain any information which either contradicts $\mathcal{K}$ or is not supported by $\mathcal{K}$. 

\rparagraph{Behavioural Tuning for Faithfulness} 
An effective model for faithful information-seeking dialogue needs to perform two actions correctly: \textbf{1)} select the correct part of information provided in  $\mathcal{K}$ (termed \textit{selectivity}) and \textbf{2)} provide the response, with the requirement to (i) inform the user when $\mathcal{K}$ contains no information relevant to $u$, or (ii) ask for clarification (termed \textit{(response) adequacy});\footnote{Put simply, in our setup response adequacy discerns between \textbf{1)} the case when the model does have the correct information in the knowledge source and should provide it versus \textbf{2)} the case when the model is certain that it cannot provide a correct answer to the user query or it does not even understand the query and requires further clarification to be able to react in the next turn. There might be other, finer-grained options of response adequacy beyond the two simple cases investigated here, but we leave those investigations to future research.} see Figure~\ref{fig: dialogue example} again. \approach aims to improve on both desiderata via behavioural fine-tuning \cite{ruder2021lmfine-tuning} of any instruction-tuned LLM. 

To instill the capability for information-seeking dialogue into the model, we perform behavioural tuning on the combination of (i) conversational QA and (ii) information-seeking dialogue datasets. In both tasks, the response has to be generated based on some knowledge source $\mathcal{K}$, making them suitable for faithful response generation. 
Further, beyond tuning on related tasks, we propose to augment the datasets to steer the model towards the {selectivity} and {adequacy} behaviour, as follows. 

For {selectivity}, ground truth $\mathcal{K}$ provided in the dataset is extended with additional knowledge sources $\mathcal{K}'$ which are irrelevant to user query $u$, serving as negative examples or distractors. Intuitively, distractors mimic the presence of information irrelevant to $u$ in $\mathcal{K}'$, this way promoting the model's selectivity. We augment ground truth knowledge source $\mathcal{K}$ with $n$ distractors; they are randomly sampled from the knowledge base of the corresponding dataset. 

For response adequacy, we augment the fine-tuning datasets with dialogues without any relevant $\mathcal{K}$ provided, making them unanswerable for the system. To construct such dialogs, for a dialogue history $\mathcal{H}$ and a corresponding user query $u$ we randomly sample unrelated knowledge sources $\mathcal{K}'$. During fine-tuning, the response $r$ is substituted with a special response  signifying that the combination of $\mathcal{H}$ and $u$ cannot be answered based on provided $\mathcal{K}'$. In our experiments, we augment the original dataset with 10\% unanswerable dialogues. 

\rparagraph{Further Task-Specific Fine-Tuning} The output of the `general' behavioural fine-tuning step is a `behaviour-specialised' LLM for factually faithful information seeking dialogue. It can be used directly `as is`, or as a starting point for further task-specific tuning, as illustrated in Figure~\ref{fig:setups}.





\section{Experimental Setup}
\label{sec:exp}

\sparagraph{Training Setup} In order to leverage inductive biases of instruction-tuned models, the input for \approach includes the following: (i) instructions to respond as factually accurately as possible, (ii) augmented knowledge source which includes: ground truth $\mathcal{K}$ and $n=4$ distractors $\mathcal{K}'$ for `answerable' dialogues, and 5 randomly sampled $\mathcal{K}'$-s for unanswerable dialogues and (iii) dialogue history which combines all the previous turns (the set $\mathcal{H}$) and the current user query $u$. An example input and instruction text are shown in Appendix~\ref{sec:input example}. The models are then trained in a standard sequence-to-sequence fashion with cross-entropy loss. The output is either ground truth responses for answerable dialogues, where knowledge source $\mathcal{K}$ contains the information to address user's query, or a predefined response \textit{`Could you please clarify or rephrase the query?'} if the dialogue is unanswerable. Training the models using \approach proceeds at turn level: dialogue history at every turn is used as input.   

\rparagraph{Datasets} To perform behavioural fine-tuning, we use a standard dataset for information seeking dialogue, FaithDial~\cite{dziri-etal-2022-faithdial}, and an established conversational QA dataset, TopiOCQA~\cite{adlakha-etal-2022-topiocqa}.  Generalisation capabilities of the models after the \approach tuning are evaluated on another domain and dataset (i.e., this could be seen as `zero-shot' from the domain adaptation perspective). For this, unless explicitly stated otherwise, we rely on a multi-domain conversational QA dataset, DoQA~\cite{campos-etal-2020-doqa}. The key statistics of the datasets are in Table~\ref{tab: dataset statistics}, with further details and data analyses in Appendix~\ref{sec: dataset characteristics}.

\begin{table}[!t]
\def\arraystretch{0.95}
\centering
\resizebox{0.999\linewidth}{!}{%
\begin{tabular}{@{}lccc@{}}
\toprule
                                                                    & FaithDial                                                       & TopiOCQA                                                       & DoQA                                                                    \\ \cmidrule(lr){2-4}
Domains                                                             & \begin{tabular}[c]{@{}c@{}}Open \\ Wikipedia-based\end{tabular} & \begin{tabular}[c]{@{}c@{}}Open\\ Wikipedia-based\end{tabular} & \begin{tabular}[c]{@{}c@{}}3\\ (Cooking, Travel, \\ Music)\end{tabular} \\
\# dialogues                                                            & 4,094 / 764 / 791                                                & 3,509 / 205 / 206                                               & 1,037 / 200 / 1,200                                                       \\
\# turns                                                            & 36,809 / 6,851 / 7,101                                             & 45,450 / 2,514 / 2,502                                            & 4,612 / 911 / 5,394                                                       \\
Avg. turns                                                          & 9                                                               & 13                                                             & 4.48                                                                    \\
\begin{tabular}[c]{@{}l@{}}Avg. length \\ of questions\end{tabular} & 17.25                                                           & 6.92                                                           & 12.99                                                                   \\
\begin{tabular}[c]{@{}l@{}}Avg. length\\ of responses\end{tabular}  & 20.29                                                           & 11.38                                                          & 10.43                                                                   \\ \bottomrule
\end{tabular}
}
\caption{Overall statistics of the used dialogue datasets. The number of conversations and turns are provided for train / dev / test splits of the datasets. } \label{tab: dataset statistics}
\end{table}
\begin{figure}[!t]
    \centering
    \includegraphics[width=0.79\linewidth]{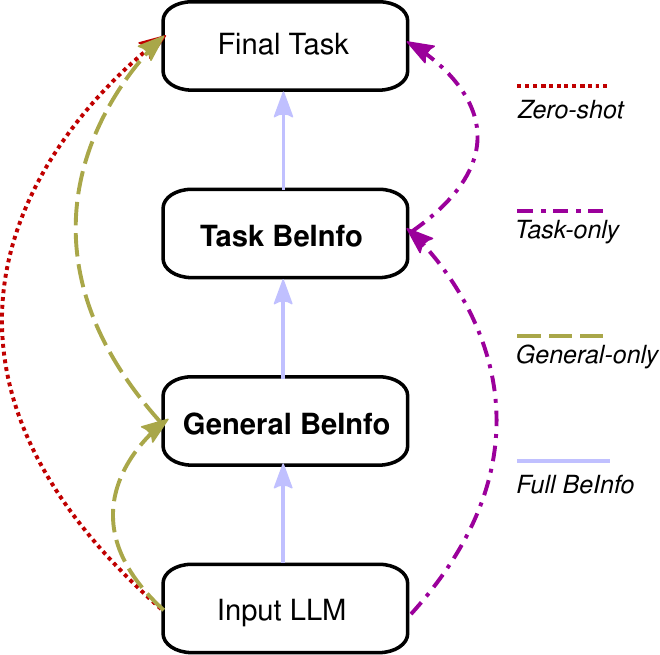}
    \caption{An overview of different fine-tuning and inference setups for LLMs with and without \approach (\S\ref{sec:exp}).}
    \vspace{-0.0mm}
    \label{fig:setups}
    \vspace{-1.5mm}
\end{figure}


\rparagraph{Models} 
Prior work~\cite{dziri-etal-2022-faithdial} has demonstrated that instruction-tuned models such as the Flan series~\cite{chung2022flan} are a very strong baseline for factuality in information-seeking dialogue. Thus, we use them as a base for the proposed method.\footnote{We again note that \approach can be applied on top of any generative model.} In the experiments, we use Flan-T5 \cite{chung2022flan} (\textsc{Base, Large} and \textsc{XL}) and Tk-Instruct-\textsc{3B} \citep{wang-etal-2022-super}. All the backbone models were pretrained on a large number of tasks with instructions, which yields faster specialisation of the models to information-seeking dialogue, especially when, as in our setup, the input/prompt includes a short description of the task.

\rparagraph{Fine-Tuning and Inference Setups}
The LLMs can be used directly in the final task in a fully \textit{zero-shot} manner or via in-context learning as `black boxes': this is a typical usage of very large models in dialogue tasks. We can also conduct \approach tuning of `smaller LLMs' via different regimes: (i) fine-tuning directly on the task data but with augmented knowledge sources (if available) (i.e., \textit{task-only} \approach); (ii) fine-tuning only on the available data from other dialogue datasets and porting the tuned model to the task in a zero-shot fashion (i.e., \textit{general-only} \approach - an example is tuning on FaithDial and TopiOCQA and using the model for DoQA, or vice versa); (iii) finally, we can run a stage of general \approach followed by in-task \approach (termed \textit{full} \approach). An overview of the different setups is provided in Figure~\ref{fig:setups}.


\rparagraph{Evaluation Metrics} We rely on automated metrics to measure \textit{lexical similarity} of the generated responses and ground truth responses: BLEU~\citep{papineni2002bleu} and ROUGE~\citep{lin-2004-rouge}. To measure \textit{semantic similarity} between generated and gold responses, we use BERTScore \citep{zhang2019bertscore}.\footnote{Similarly to \citet{daheim2023elasticwr}, we use \textit{deberta-large-mnli} as an underlying model for computing the score.} To evaluate \textit{faithfulness}, we use BERTScore and token-level precision between the generated response and the knowledge source $\mathcal{K}$. We denote BERTScore between ground truth and generated responses as ``BERTS'' and one between the knowledge source $\mathcal{K}$ and generated responses as  ``$\mathcal{K}$-BERTS''. In both cases we use BERTScore-F1. Token-level precision between the generated response and knowledge source $\mathcal{K}$ ($\mathcal{K}$-Precision; \citeauthor{adlakha2023evaluating}, \citeyear{adlakha2023evaluating}) measures the proportion of tokens in generated response which occur in $\mathcal{K}$. Prior work \cite{adlakha2023evaluating} demonstrates that $\mathcal{K}$-Precision has the highest correlation with human (as well as GPT-4-elicited) faithfulness judgements among different automated metrics.  

\rparagraph{Hyperparameters and Training Details} 
\approach was implemented using HuggingFace Transformers library~\cite{wolf-etal-2020-transformers}. The models were trained with AdamW~\citep{Loschilov:2018iclr}. With \approach, we tune for 5 epochs, the learning rate is $5$e-$5$; when tuning the model to a specific dataset, we run it for 10 epochs with the learning rate of $5$e-$6$. We use the warm-up rate of 0.1 and linear decay, with the default weight decay rate of 0.01. Beam search is run with the beam size of 10.


\section{Results and Discussion}
\paragraph{Faithfulness on Unseen Data.} One of the main aims of behavioural fine-tuning with \approach is to increase the factual faithfulness of responses in zero-shot domain transfer, on unseen data in any domain. Therefore, we start by presenting the results of the variant tuned with \approach on FaithDial plus TopiOCQA, where inference is run on the dataset unseen during \approach tuning: DoQA (i.e., general \approach from Figure~\ref{fig:setups}). The results are presented in Table~\ref{tab: results unseen doqa}. They confirm that \approach substantially improves faithfulness while either improving or only minimally affecting the similarity between generated responses and the gold response. Importantly, the improvements hold across different model sizes: Flan-T5 \textsc{Base}, \textsc{Large} and \textsc{XL} with 250M, 780M and 3B parameters, respectively.

\begin{table}[]
\def\arraystretch{0.85}
\centering
\resizebox{0.99\linewidth}{!}{%
\begin{tabular}{@{}lrrrrr@{}}
\toprule
Model        & \multicolumn{1}{l}{\mustardcell BLEU} & \multicolumn{1}{l}{\mustardcell ROUGE} & \multicolumn{1}{l}{\mustardcell BERTS} & \multicolumn{1}{l}{\peachcell$\mathcal{K}$-BERTS} & \multicolumn{1}{l}{\peachcell$\mathcal{K}$-Precision} \\ \midrule
Flan-T5$_{\textsc{Base}}$ & 22.89                                            & 34.46                                             & 61.60                                                 & 67.75                                                   & 90                                                    \\
~~~+\approach    & 22.76                                            & 34.04                                             & 61.71                                                 & 77.55                                                   & 100                                                    \\ \cmidrule{1-6}
Flan-T5$_{\textsc{Large}}$    & 26.16                                            & 39.57                                             & 64.61                                                 & 71.38                                                   & 93.86                                                  \\
~~~+\approach    & 26.34                                            & 38.55                                             & 63.19                                                 & 75.55                                                   & 100                                                     \\ \cmidrule{1-6}
Flan-T5$_{\textsc{XL}}$   & 28.66                                            & 41.99                                             & 65.89                                                 & 67.21                                                   & 94.12                                                   \\
~~~+\approach    & 26.65                                            & 39.39                                             & 64.60                                                 & 80.19                                                   & 100                                                       \\ \bottomrule
\end{tabular}
}
\vspace{-0.5mm}
\caption{Results on DoQA without any in-task \approach tuning. The models are tuned on a combination of FaithDial and TopiOCQA. The results are averaged across three domains in DoQA -- \textit{Cooking, Travel and Movies}. Full results are presented in Appendix~\ref{app: doqa per domain}.}
\label{tab: results unseen doqa}
\vspace{-1mm}
\end{table}

\rparagraph{Using a Smaller Dataset for \approach Tuning} The previous results from Table~\ref{tab: results unseen doqa} show \approach's effectiveness when tuned on two reasonably sized datasets, FaithDial with 36,809 turns, and TopiOCQA with 45,450 turns. Now, we test the opposite direction: fine-tuning \approach on a smaller-scale dataset like DoQA (4,612 turns) and evaluating zero-shot on FaithDial. Besides further testing the versatility of the approach, we also probe sample efficiency of the approach and its adaptability to smaller datasets and computational budgets. 



Results in Table \ref{tab: results faithdial pretuned doqa} suggest that tuning the models with \approach even on smaller datasets without any subsequent in-task tuning consistently improves the factuality of generated responses. Especially large gains were observed for larger models, both for faithfulness and semantic similarity between the generated responses and the ground truth, indicating the potential for sample efficiency of \approach.  Similar trends were observed when evaluating on TopOCQA instead of FaithDial; see Appendix~\ref{sec: smaller dataset topiocqa}.

\begin{table}[]
\def\arraystretch{0.85}
\centering
\resizebox{0.99\linewidth}{!}{%
\begin{tabular}{@{}lrrrrr@{}}
\toprule
Model        & \multicolumn{1}{l}{\mustardcell BLEU} & \multicolumn{1}{l}{\mustardcell ROUGE} & \multicolumn{1}{l}{\mustardcell BERTS} & \multicolumn{1}{l}{\peachcell $\mathcal{K}$-BERTS} & \multicolumn{1}{l}{\peachcell $\mathcal{K}$-Precision} \\ \midrule
Flan-T5$_{\textsc{Base}}$ & 4.15	& 19.5	& 53.78	& 42.17	& 0                                                    \\
~~~+\approach    & 5.39	& 21.04 &	54.68 &	70.03 &	27.78                                         \\ \cmidrule{1-6}
Flan-T5$_{\textsc{Large}}$    & 5.01 &	20.02	& 54.56	& 61.77	& 0                                                  \\
~~~+\approach    & 9.27	& 29.29	& 61.75 &	86.58 &	6.67                                                     \\ \cmidrule{1-6}
Flan-T5$_{\textsc{XL}}$   & 5.26 &	22.21	& 56.13 &	65.52	& 6.67                                                   \\
~~~+\approach    & 10.2	& 30.76	& 62.78	& 88.50 & 100                                                       \\ \bottomrule
\end{tabular}
}
\vspace{-0.5mm}
\caption{Zero-shot results on FaithDial. The models are tuned on DoQA.}
\label{tab: results faithdial pretuned doqa}
\vspace{-1.5mm}
\end{table}


\rparagraph{Different Instruction-Tuned Models} Previous results have already verified that \approach can be applied to Flan models of different sizes, and we now evaluate its impact on another instruction-based model: Tk-Instruct-\textsc{3B}. We fine-tune the models again on FaithDial and TopiOCQA and evaluate their performance on DoQA's \textit{Travel} domain test set. While the absolute scores, as expected, do differ between different underlying models, the results in Table \ref{tab: results different instruction tuned models} indicate the positive effect of \approach also on Tk-Instruct-\textsc{3B}.

\begin{table}[]
\def\arraystretch{0.85}
\centering
\resizebox{0.99\linewidth}{!}{%
\begin{tabular}{@{}lrrrrr@{}}
\toprule
Model        & \multicolumn{1}{l}{\mustardcell BLEU} & \multicolumn{1}{l}{\mustardcell ROUGE} & \multicolumn{1}{l}{\mustardcell BERTS} & \multicolumn{1}{l}{\peachcell$\mathcal{K}$-BERTS} & \multicolumn{1}{l}{\peachcell$\mathcal{K}$-Precision} \\ \midrule
Flan-T5$_\textsc{XL}$ & 25.88	& 41.68 &	66.91	& 66.42 & 100	                                                 \\
~~~+\approach    & 23.28	& 36.22 & 63.02	& 81.77 & 100	                                                   \\
Tk-Instruct-\textsc{3B} & 20.23	& 31.60	& 58.47	& 69.45 & 100	                                                 \\
~~~+\approach    & 29.19	& 42.56	& 66.24	& 	70.58  & 97.8                                              \\ \bottomrule
\end{tabular}
}
\vspace{-0.5mm}
\caption{Zero-shot results on DoQA \textit{Travel} domain. The models are tuned on FaithDial + TopiOCQA.}
\label{tab: results different instruction tuned models}
\vspace{-1mm}
\end{table}

\rparagraph{\approach with Task-Specific Fine-Tuning} We have demonstrated that the models tuned with \approach largely improve factual faithfulness on unseen datasets and domains (i.e., the general \approach setup). Here, we study whether these models can serve as an effective starting point for continued task-specific fine-tuning. To this end, we first tune the models with \approach on the combination of FaithDial and TopiOCQA as before, and then continue fine-tuning/specialising the model on a single dataset (e.g., FaithDial or TopiOCQA): the \textit{full} setup from Figure~\ref{fig:setups}.\footnote{We present the results only with Flan-T5$_{\textsc{BASE}}$ as preliminary experiments with larger model sizes demonstrated similar relative trends.}

\begin{figure}[!t]
    \centering
    \includegraphics[width=0.4\textwidth]{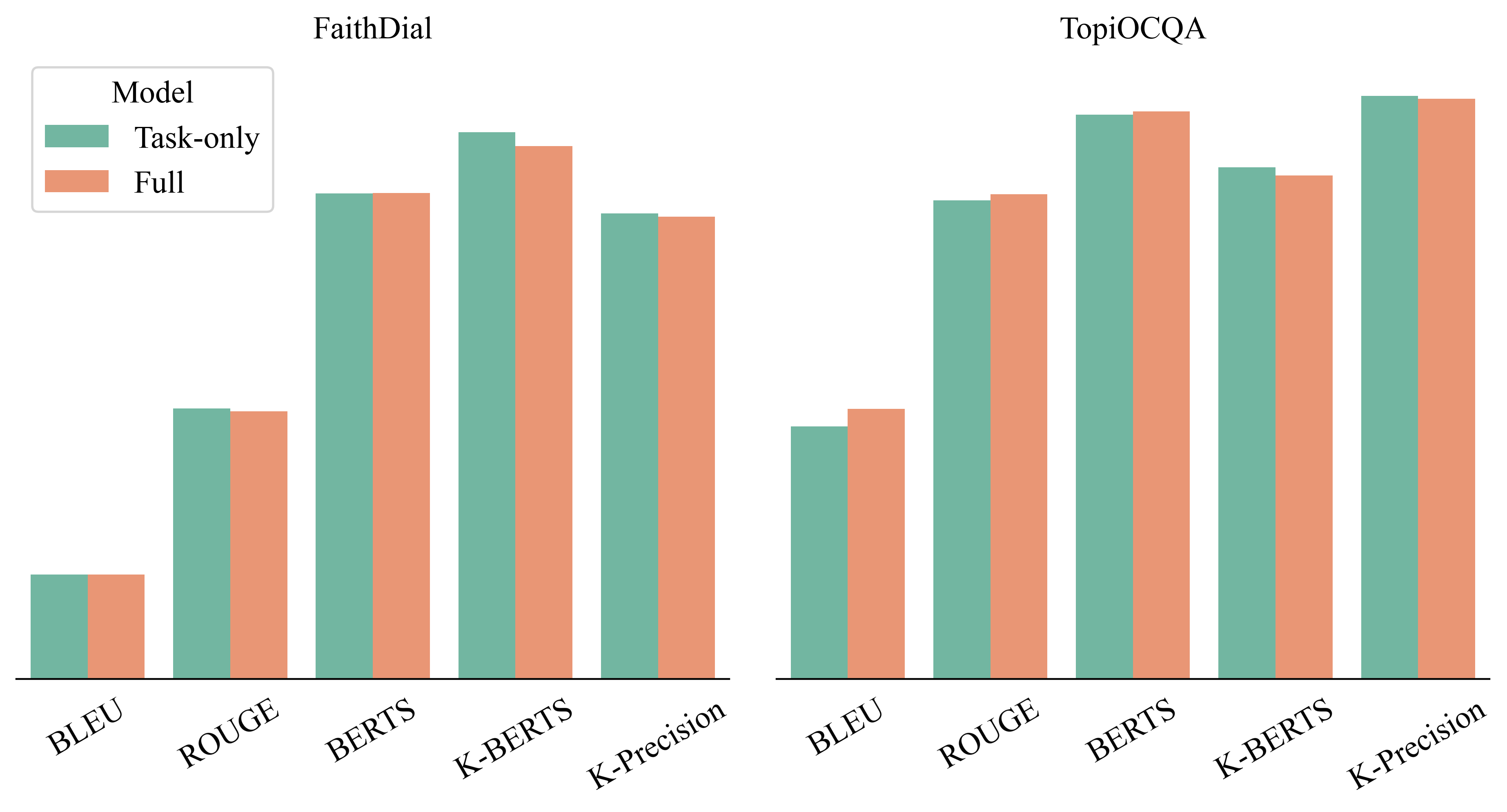}
    \vspace{-1mm}
    \caption{Results of task-specific tuning on FaithDial (left) and TopiOCQA (right). \textit{`Task-only'} denotes Flan-T5 tuned directly on FaithDial or TopiOCQA, again with knowledge distractors. \textit{`Full'} denotes the model first tuned with \approach on both datasets and then further tuned on each of the datasets; see Figure~\ref{fig:setups}.}
    \label{fig: results task specific tuning}
    \vspace{-2mm}
\end{figure}

Figure~\ref{fig: results task specific tuning} demonstrates that already \textit{task-only} \approach yields strong performance, while models with \approach perform on par or better on average than the models which were tuned to a specific dataset \textit{both} on semantic similarity of generated responses and factual faithfulness. While prior work~\cite{daheim2023elasticwr} typically optimised one aspect (e.g., semantic similarity) at the expense of the other (faithfulness), and vice versa, here we show that through the use of knowledge distractors \approach achieves competitive performance on both aspects and retains the cross-dataset generalisation ability.

\begin{table}[!t]
\def\arraystretch{0.9}
\centering
\resizebox{0.99\linewidth}{!}{%
\begin{tabular}{@{}lrrrrr@{}}
\toprule
Model        & \multicolumn{1}{l}{\mustardcell BLEU} & \multicolumn{1}{l}{\mustardcell ROUGE} & \multicolumn{1}{l}{\mustardcell BERTS} & \multicolumn{1}{l}{\peachcell $\mathcal{K}$-BERTS} & \multicolumn{1}{l}{\peachcell $\mathcal{K}$-Precision} \\ \midrule
Task-only: FaithDial & 27.29 &	41.71 &	75.31 &	64.80	& 73.38     \\
General-only & 38.75  &	69.57 &	80.40	& 72.24  &	81.24  \\
Full: FaithDial & 33.87 &	55.85 &	78.46	& 74.18 &	79.63  \\
\hdashline
\rowcolor{Gray} Task-only: TopiOCQA & 36.24	& 68.64 &	80.94 &	73.38 &	83.63  \\\bottomrule
\end{tabular}
}
\vspace{-0.5mm}
\caption{Results on TopiOCQA when the \approach model is further fine-tuned on FaithDial after the original FaithDial + TopiOCQA fine-tuning. `Task-only: TopiOCQA' denotes direct tuning on TopiOCQA, which serves as an upper bound in this experiment.}
\label{tab: results cross dataset topiocqa}
\vspace{-2mm}
\end{table}


\rparagraph{\approach versus Catastrophic Forgetting}
Further, one issue which might arise from further specialising a model to a given task/dataset is a well-known phenomenon of \textit{catastrophic forgetting}: pretrained language models are prone to forgetting previously learnt knowledge or skills when tuned on new data~\cite{de-cao-etal-2021-editing, yuan-etal-2023-pretrained}. To evaluate whether the models would retain their ability to respond faithfully to examples considerably different from the ones seen during fine-tuning, we evaluate the models tuned on FaithDial on TopiOCQA.\footnote{We focus on TopiOCQA as the true responses in the dataset are more grounded in the knowledge source $\mathcal{K}$ (see Appendix~\ref{sec: dataset characteristics}).}  The scores in Table~\ref{tab: results cross dataset topiocqa} demonstrate that even after continued fine-tuning on FaithDial the model retains high faithfulness scores  on TopiOCQA (cf. $\mathcal{K}$-BERTS and $\mathcal{K}$-Precision). At the same time, degradation in scores for similarity to ground truth responses shows that further tuning largely influences the style/form of the responses. The average response length in FaithDial in considerably larger than that in TopiOCQA (see Appendix \ref{sec: dataset characteristics}), meaning that further tuning on FaithDial leads the model to generate longer responses not matching the gold responses in TopiOCQA. In other words, these results show that further fine-tuning might influence the surface form of the responses but not the desired skill to respond faithfully gained with \approach. In practice, a general model tuned with \approach on a wide range of tasks/domains and then specialised to one of them would still retain its ability to respond faithfully for \textit{any} of the domains seen in the general `behavioural tuning' step.

\section{Evaluating \approach on Real Conversations} 
\label{s:gn}
\sparagraph{Experimental Setup}
To probe the potential of \approach for boosting real user-facing production systems, we rely on a small internal dataset of 200 fully anonymised dialogs with real users in the hotel reservation domain (termed \textsc{hotel-200} henceforth); the dialogues concern hotel bookings and FAQ-s about its various facilities. It is crucial to evaluate the models on examples also collected from real user-system communication, as the language use is considerably different to some established datasets such as DoQA or FaithDial compiled via crowdsourcing work. For instance, the average length of the user query in \textsc{hotel-200} is only 6.35 tokens, while it is 17.25 in FaithDial or 13 in DoQA (cf., Table~\ref{tab: dataset statistics}). 

As the data comes from real conversations, there are no gold responses which could be used for automated evaluation. Thus, we resort to evaluation of correctness/factual faithfulness with an LLM: here, we use GPT4 (termed \textit{GPT4-Eval} henceforth) as its judgements were shown to be most correlated with human judgements \cite{adlakha2023evaluating}.\footnote{As running evaluation with large models such as GPT4 behind proprietary APIs incurs large costs~\citep{adlakha2023evaluating}, we only evaluate the outputs for a smaller dataset where other means of evaluation cannot be used.} For \textit{GPT4-Eval} we prompt GPT4 to act as the evaluator providing it with natural language instructions, knowledge source $\mathcal{K}$, conversation history $\mathcal{H}$ with user query $u$ and the system-generated response. In the instructions we request the model to rate generated responses on a 7-point Likert-scale for faithfulness, available in Appendix~\ref{app:likert}.

We compare the following models and their configurations: (i) GPT4 itself as the model responding to user query $u$, (ii) Falcon-40B~\cite{falcon40b} as a strong open-source LLM,\footnote{Falcon-40B was an open-source large language model with state-of-the-art results at the time of the experimentation.} (iii) Flan-T5$_{\textsc{XL}}$ tuned with \approach, under the three different regimes illustrated before in Figure~\ref{fig:setups} (\textit{general-only, task-only, full}). For the \textit{general-only} and the first stage of the \textit{full} \approach, we again rely on the combination of FaithDial and TopiOCQA datasets.

To obtain data for the task-specific tuning stage, we collect 2,000 examples from the same conversational system, then generate `silver' responses via GPT4 and treat the silver responses as true outputs for task-specific fine-tuning.\footnote{Note that here we use the GPT4 model for three different purposes: (i) as an evaluator; (ii) as an actual baseline system; (iii) as a `silver data generator'.} 

\rparagraph{Results and Discussion}
The main results are reported in Table~\ref{tab: hidden dataset results}. While the zero-shot \approach approach with Flan-T5$_{\textsc{XL}}$ achieves a reasonably high average faithfulness score in absolute terms, it is still far from that of GPT4, which serves as an upper bound zero-shot system. Most importantly, the progress in scores reveals the importance of various \approach fine-tuning stages. Even the \textit{general-only} fine-tuning stage without seeing a single in-domain training example yields an average score which is substantially higher than that of the original Flan-T5$_{\textsc{XL}}$ as well as higher than the score obtained by the 40B Falcon model. Further, the scores indicate the importance of being able to fine-tune smaller models with in-domain data: the 3B model tuned with the full \approach even outperforms GPT4 on \textit{GPT4-Eval}, and it also obtains strong performance with \textit{task-only} \approach. 

These results further support our hypothesis that \approach actually `behaviourally prepares' the models to respond to user's queries in a factually faithful manner and tuning on further task-specific data only amplifies its impact as it gets further adapted to the domain. Put simply, behavioural fine-tuning via \approach performs structural (or behavioural) adaptation, while further task-specific fine-tuning combines the behavioural adaptation with (semantic) domain adaptation.

\begin{table}[]
\def\arraystretch{0.95}
\centering
\resizebox{0.999\linewidth}{!}{%
\begin{tabular}{@{}cccccc@{}}
\toprule
\rowcolor{Gray}                   GPT-4 & Falcon-40B & XL-original & XL+\approach (g) & XL+\approach (t) & XL+\approach (f) \\\midrule
4.63  & 3.60 & 3.55 &  3.98 &  4.46  & \textbf{4.81}            \\ \bottomrule
\end{tabular}
}
\vspace{-0.5mm}
\caption{Averaged \textit{GPT4-Eval} scores (higher is better) on the \textsc{hotel-200} dataset. XL denotes the Flan-T5$_{\textsc{XL}}$ model taken-off-the-shelf (XL-original) or fine-tuned via three different regimes of \approach (t=task-only; g=general-only; f=full).}
\label{tab: hidden dataset results}
\vspace{-1.5mm}
\end{table}
\begin{figure*}[!t]
    \centering
    \includegraphics[width=0.959\textwidth]{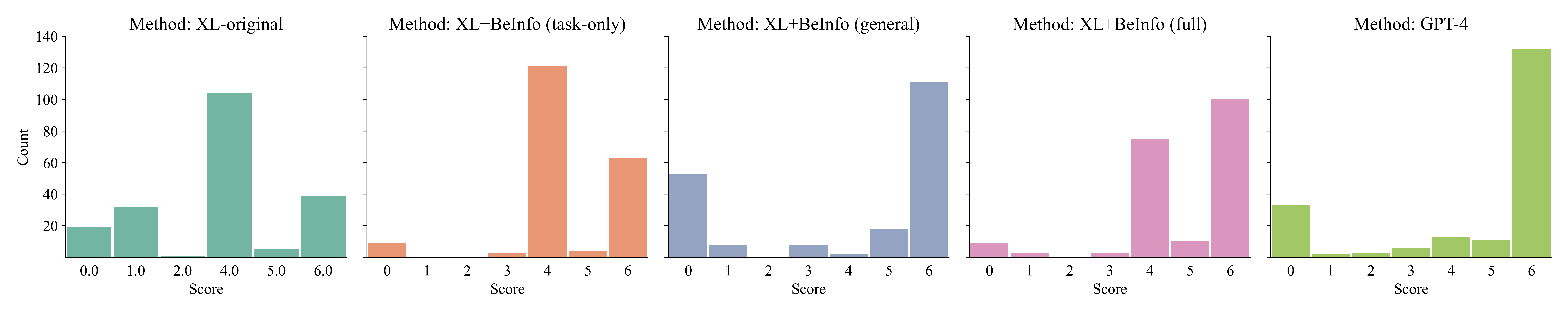}
    \vspace{-2mm}
    \caption{Distribution of \textit{GPT4-Eval} scores of 4 variants based on Flan-T5$_{\textsc{XL}}$ and GPT-4. See Table~\ref{tab:likert} in Appendix~\ref{app:likert} for the interpretation of the individual scores.}
    \label{fig:discribution_scores_tuned}
    \vspace{-2mm}
\end{figure*}


\rparagraph{Ablation: Distributions of Scores} 
We further study the actual distributions of \textit{GPT4-Eval} scores for the four models variants of Flan-T5$_{\textsc{XL}}$ and compare it against the distribution obtained by GPT-4. The distributions are shown in Figure~\ref{fig:discribution_scores_tuned}. As only a small fraction of responses is labelled with intermediate scores ({1,2,3,5}),  the core differences lie in relative distribution of \textit{perfect}, \textit{poor} and \textit{`not great, not terrible'} responses (scores 6,0 and 4, respectively).\footnote{Score 4 usually corresponds to the system responding with a generic clarification question or notifying the user that the information is not available.} The model tuned with task-only \approach rarely provides wrong facts but mostly responds with \textit{not great, not terrible} responses which do not mislead the user but might not be helpful. On the other hand, the model tuned with the general-only \approach and GPT-4 both yield responses that typically fall into the extreme categories. In other words, the responses are either perfect (score 6) or will provide the user with wrong information (score 0), which is not desirable for a user-facing system.  The model tuned with the full \approach combines the benefits of behavioural tuning with the use of in-domain data: the model produces the least factually unfaithful responses (score 0) while maintaining the ability to respond with information relevant to the user's query (a large number of responses with scores 6). In sum, `pre-tuning' the model with the general-only \approach stages raises faithfulness of the model by extracting relevant information from the knowledge source $\mathcal{K}$ while further tuning on task-specific data further helps avoid providing misleading or irrelevant information to the user.


\rparagraph{Faithfulness versus Abstractiveness}  Increasing faithfulness of a model to the underlying knowledge source $\mathcal{K}$ can lead the model to respond with large \textit{extracted spans} of text from $\mathcal{K}$. Ideally, the responses should be abstractive but factually faithful: in other words, they should transmit the information provided in the knowledge source $\mathcal{K}$ but use different means of expression of it. As in prior work \cite{dziri-etal-2022-faithdial, daheim2023elasticwr}, we use the Density metric proposed by \citet{grusky-etal-2018-newsroom} to measure abstractiveness. This measures average length of spans copied from the knowledge source. We focus on $\mathcal{K}$-BERTScore to measure faithfulness in this experiment: the average length of the knowledge source $\mathcal{K}$ ($\approx$120 words on average) is relatively large with respect to the length of generated responses ($\approx$ 18--25 words) making $\mathcal{K}$-BERTScore suitable for this case. 

\begin{figure}[!t]
    \centering
    \includegraphics[width=0.39\textwidth]{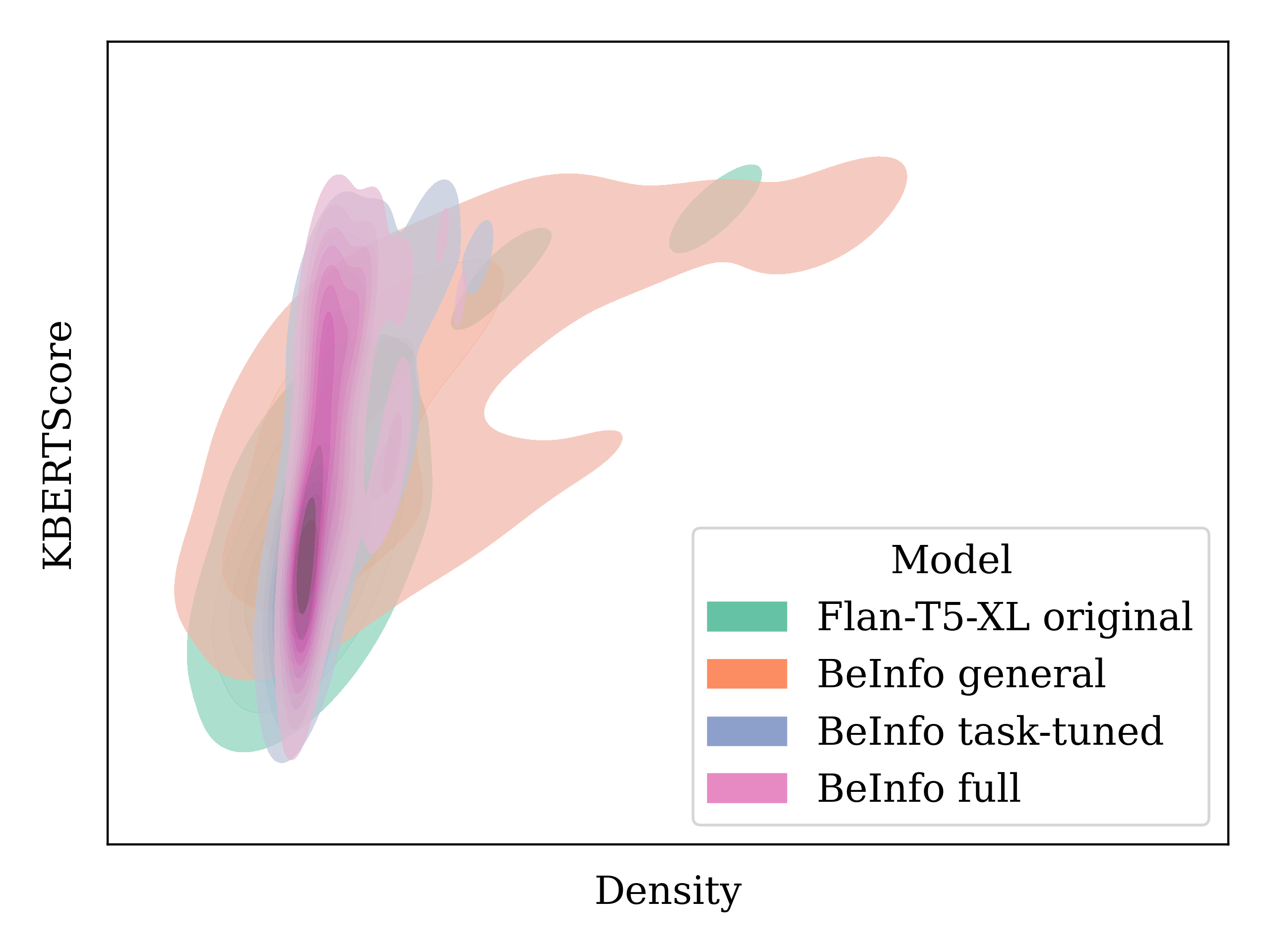}
    \vspace{-0.5mm}
    \caption{Density and $\mathcal{K}$-BERTScore on \textsc{hotel-200} illustrating the trade-off between faithfulness (y-axis) and abstractiveness (x-axis) for Flan-T5$_{\textsc{XL}}$ for different setups: (i) XL-original: `off-the-shelf' Flan-T5$_{\textsc{XL}}$; (ii) \approach general-only: Flan-T5$_{\textsc{XL}}$ tuned with \approach on FaithDial and TopiOCQA without any in-task data; iii) \approach task-only:  Flan-T5$_{\textsc{XL}}$ finetuned only on task-specific data; iv)  \approach full. Numeric results are provided in Appendix \ref{sec:results_faithful_abstractiveness}.}
    \label{fig:faithfulness_vs_abstractiveness}
    \vspace{-1mm}
\end{figure}

Figure~\ref{fig:faithfulness_vs_abstractiveness} illustrates the trade-off between faithfulness and abstractiveness for Flan-T5$_{XL}$ under different fine-tuning setups on the \textsc{hotel-200} dataset. The results demonstrate that general-only fine-tuning with \approach improves the model's factuality but increases the extractiveness of the responses. Tuning on task-specific data helps to raise the abstractiveness of the responses. Further analyses and comparisons (cf. results in Appendix~ \ref{sec:results_faithful_abstractiveness}) demonstrate that Flan-T5$_{\textsc{XL}}$ tuned with the full \approach is on par with GPT-4 and better than a considerably larger Falcon-40B model.


\rparagraph{Human Evaluation} In addition to automatic metrics, we also conduct human evaluation on \textsc{hotel}-200 with two annotators. They were tasked to rate each response on factuality using the same Likert-scale as used for \textit{GPT4-Eval} (see Appendix~\ref{app:likert}).  Three models were assessed: XL+\approach-general, XL+\approach-task-specific and XL+\approach-full. Average human factuality scores were 3.64, 4.62 and 4.95, respectively. This further proves the effectiveness of behavioural tuning for improved factuality. To further assess relevance of automatic \textit{GPT4-Eval}, we also compute Pearson's correlation coefficient $\rho$ between human judgements and GPT4-Eval scores. This results in strong positive correlation with $\rho = 0.52$, indicating that \textit{GPT4-Eval} can be used as a reasonable automatic proxy.

\section{Related Work}
\sparagraphnodot{Mitigating Hallucinations in Information-Seeking Dialogue} has achieved increased interest recently with the omnipresence of large language models \cite{wang-etal-2023-self-instruct, chuang2023dola, daheim-etal-2022-controllable, daheim2023elasticwr, zhang2023-sirens-in-hallucinations}. Previous methods can be largely divided into those which increase factuality of pretrained models via further training or modification of the generation procedure. The former includes, e.g., tuning the models with contrastive learning \citep{sun2023contrastive} or a special \textit{focus learning} loss which reduces hallucinations on token level \cite{deng-etal-2023-towards}. The latter includes, e.g., conditioning generation process on special control tokens \citep{rashkin-etal-2021-increasing}, task arithmetic \citep{daheim2023elasticwr} or training a critic network which can detect problematic tokens and replace them~\citep{dziri-etal-2021-neural}. Other approaches have been developed to specifically improve faithfulness with respect to retrieved knowledge source in decoding. One proposed option is to do context-aware decoding (CAD; \citeauthor{shi2023trusting}, \citeyear{shi2023trusting}) where generative probabilities are contrasted between those based only on user query and those based on the user query \textit{and} the knowledge source. The aim is to force LLMs to rely more on the knowledge source than the model's internal knowledge from pretraining. In contrast to CAD, \citet{chuang2023dola} propose to contrast generation probabilities from different layers of LLMs to promote factual knowledge in the resulting output probabilities.  

\rparagraph{Improving Faithfulness via Supervised Tuning} Task-specific supervised fine-tuning could be seen as an option to improve faithfulness of the model's responses \cite{zhang2023-sirens-in-hallucinations}. Prior work \cite{cao2023instruction-mining, chen2023alpagasus} has demonstrated that fine-tuning on higher-quality data improves the model's factuality on benchmarks such as TruthfulQA \cite{lin-etal-2022-truthfulqa}. In contrast, supervised fine-tuning on the data which includes numerous irrelevant or factually inconsistent responses can lead the model to amplifying the noise in the training data. A recent analysis from \citet{dziri-etal-2022-origin} has shown that over 60\% of responses in three standard datasets for information-seeking dialogue (WoW, \citeauthor{dinan2018wizardofwikipedia}, \citeyear{dinan2018wizardofwikipedia}; CMU-DoG, \citeauthor{zhou-etal-2018-dataset}, \citeyear{zhou-etal-2018-dataset}; and TopicalCHAT, \citeauthor{Gopalakrishnan2019topical-chat}, \citeyear{Gopalakrishnan2019topical-chat}) contain hallucinations, making them unsuitable for supervised fine-tuning aimed at improving factuality. To resolve this, \citet{dziri-etal-2022-faithdial} released a corrected version of WoW where the responses were fixed to be factually consistent with the knowledge source. As behavioural fine-tuning heavily relies on the quality of the underlying data, we have carefully selected and resorted to FaithDial and TopiOCQA in the first stage of \approach with highest factual faithfulness of their ground truth responses (see Appendix~\ref{sec: dataset characteristics} for further details).

\section{Conclusion and Future Work}

We presented \approach, a simple yet effective method that applies behavioural fine-tuning of large language models underlying information-seeking dialogue systems, with the goal of improving factuality of system responses. Instruction-tuned models are fine-tuned on a collection of publicly available dialogue data for two related tasks, conversational question answering and information-seeking dialogue, where the model must use the correct knowledge source among several `knowledge distractors' and provide a factually correct and adequate response. The main results indicated the effectiveness of \approach both in in- and cross-dataset setups. In addition, we demonstrated that further tuning on task-specific data might yield further gains in terms of faithfulness as well as reducing extractiveness, also in experiments with real conversations from a production-ready dialogue system. 

This work leads up to several potential directions of future work. Firstly, \approach is orthogonal to other existing approaches to improving faithfulness. For instance, a combination of CAD \citep{shi2023trusting} and \approach could further improve factuality of responses. Secondly, \approach was evaluated on information-seeking dialogue. Another interesting direction could be to applying it to other language generation tasks where faithfulness to the knowledge sources is crucial, such as summarisation. Furthermore, the effectiveness of the approach can be also tested on other instruction-tuned models (e.g., T0, \citeauthor{sanh2021multitask}, \citeyear{sanh2021multitask}) and models of larger sizes, e.g., Flan-UL2 and beyond.\footnote{Due to a large number of experiments coupled with computational constraints and feasibility, we focus on models that do not go beyond 3B parameters.}

The code and models will be made available online at \url{[URL]}, allowing the research community to build stronger models for factually faithful information-seeking dialogue.

\section*{Limitations}
The experiments could be further extended by altering how the knowledge distractors $\mathcal{K}'$ are sourced. Firstly, the impact of the number $n$ of knowledge distractors $\mathcal{K}'$ on faithfulness performance should be further studied. Also, another extension on this front concerns different heuristics of how $\mathcal{K}'$ is sampled. Namely, in our experiments they were sampled at random, while getting $\mathcal{K}'$ which are semantically similar or distant from the true knowledge source $\mathcal{K}$ or user query $u$ might further impact performance. 

In the experiments we focus on three widely used datasets for information seeking dialogue and two instruction-tuned models. \approach can be further extended to other datasets such as CoQA \cite{reddy-etal-2019-coqa}, MultiDoc2Dial \cite{feng-etal-2021-multidoc2dial} or the DSTC9 \cite{kim-etal-2020-beyond} extension of MultiWOZ 2.1 \cite{eric-etal-2020-multiwoz}. The evaluation on production-ready dialogues, due to associated costs of evaluation, is conducted on 200 dialogues, and we plan to run a larger-scale analysis, also spanning other dialogue domains, in future work.


We also tested whether \approach can be used with parameter-efficient finetuning (PEFT) to reduce its computational cost. Our preliminary experiments proved that \approach can be effectively combined with PEFT. However, as PEFT techniques are out of the scope of the paper and their use is orthogonal to the main experiments reported in this work, we leave out the preliminary results and focus on full fine-tuning as our main setup. 

Given that \approach uses instruction-tuned models and `behaviourally' tunes them with a predefined instruction, additional experimentation could be conducted on how the wording of the instruction influences the performance and whether one can induce higher factuality by just changing the instruction text. 

Finally, the work on improving knowledge retrieval systems as done e.g. by \citet{kdd2023} is out of scope of this work, and we focus on reducing hallucinations of LLMs in information-seeking dialogue directly, without the intervention to the knowledge retrieval component.

\bibliography{anthology,custom}

\clearpage
\appendix
\section{Example of Input}
\label{sec:input example}

\begin{figure}[!t]
    \centering
    \includegraphics[width=0.4\textwidth]{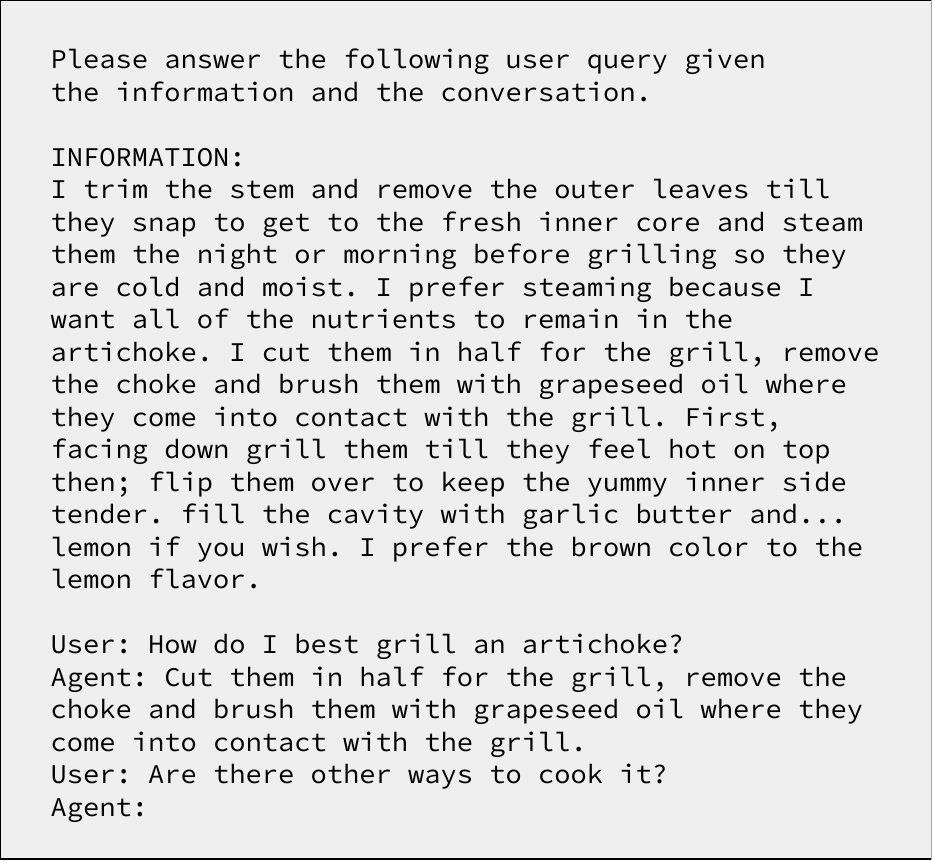}
    \caption{Example with the prompt used in \approach.}\label{fig: example instruction}
\end{figure}

An example with instructions used in \approach is shown in Figure \ref{fig: example instruction}. The used prompt is similar to the one which proved successful for conversational question-answering in \citet{adlakha2023evaluating}.

\section{Additional Dataset Statistics and Characteristics} \label{sec: dataset characteristics}
We present overall statistics of the datasets used for \approach and evaluation in Table~\ref{tab: dataset statistics}.

Additionally, we analyse the characteristics of factual faithfulness of the true responses with respect to the knowledge source. The results in Table \ref{tab: dataset characteristics} demonstrate that
the responses in FaithDial \cite{dziri-etal-2022-faithdial} are semantically most similar to their knowledge source, which is in line with the dataset collection procedure aimed to make the dataset more factual than the original responses. Similarity of contextual semantic token representations (BERTS-F1) is reversely correlated to lexical overlap between the response and knowledge source. 

As \approach is aimed at improving the model's general factual faithfulness, the results suggest that FaithDial \cite{dziri-etal-2022-faithdial} and TopiOCQA \cite{adlakha-etal-2022-topiocqa} are best used for behavioural tuning and DoQA for testing the out-of-distribution capabilities of the model. The former two have a large semantic but not literal overlap between the knowledge source and the corresponding golden response, meaning that the behavioural tuning will not lead to model learning to `\textit{copy-paste}' from the knowledge source to the response.

\begin{table}[!t]
\def\arraystretch{0.95}
\centering
\resizebox{0.9\linewidth}{!}{%
\begin{tabular}{@{}llrrr@{}}
\toprule
                        &             & \multicolumn{1}{l}{FaithDial} & \multicolumn{1}{l}{TopiOCQA} & \multicolumn{1}{l}{DoQA}   \\ \midrule
\multirow{2}{*}{($y$, $\mathcal{K}$)} & $\mathcal{K}$-BERTS-F1   & 67.31                        & 62.91                       & 52.48                     \\
                        & $\mathcal{K}$-Precision & 46.23                        & 80.67                       & 97.73                     \\ \midrule
$y$                       &      Avg. length       & 17.17    & 10.89   & 13.29\\ \bottomrule
\end{tabular}
}
\caption{BERTScore-F1 and K-Precision between the ground truth knowledge source $\mathcal{K}$ and gold response $y$. Average length is calculated as an arithmetic mean of number of whitespaced words in a response.}
\label{tab: dataset characteristics}
\end{table}

\section{Per-Domain Performance on DoQA} \label{app: doqa per domain}

Tables \ref{tab: results doqa cooking} -- \ref{tab: results doqa travel} present per-domain results of \approach (general-only) on DoQA.

\begin{table}[!ht]
\def\arraystretch{0.9}
\centering
\resizebox{0.99\linewidth}{!}{%
\begin{tabular}{@{}lrrrrr@{}}
\toprule
Model        & \multicolumn{1}{l}{\mustardcell BLEU} & \multicolumn{1}{l}{\mustardcell ROUGE} & \multicolumn{1}{l}{\mustardcell BERTS} & \multicolumn{1}{l}{\peachcell $\mathcal{K}$-BERTS} & \multicolumn{1}{l}{\peachcell $\mathcal{K}$-Precision} \\ \midrule
Flan-T5$_{\textsc{Base}}$ & 23.77 &	34.19 &	61.54	& 67.80 &    100.0	                                             \\
~~~+\approach    & 23.96	& 34.65	& 61.74	& 79.54 &    100.0	                            \\ \cmidrule{1-6}
Flan-T5$_{\textsc{Large}}$    & 27.35	& 39.79 &	64.83	& 71.78 & 100.0	                                             \\
~~~+\approach    & 28.17	& 40.57	& 64.07 &	77.77 & 100.0	                                                   \\ \cmidrule{1-6}
Flan-T5$_{\textsc{XLarge}}$   & 32.16 &	42.99	& 65.93	& 68.19 & 100.0	                                                 \\
~~~+\approach    & 28.76 &	42.68	& 65.97 &	81.65 & 100.0	                                                   \\ \bottomrule
\end{tabular}
}
\caption{Zero-shot results on DoQA \textit{Cooking}.}\label{tab: results doqa cooking}
\vspace{-1.5mm}
\end{table}

\begin{table}[!ht]
\def\arraystretch{0.9}
\centering
\resizebox{0.99\linewidth}{!}{%
\begin{tabular}{@{}lrrrrr@{}}
\toprule
Model        & \multicolumn{1}{l}{\mustardcell BLEU} & \multicolumn{1}{l}{\mustardcell ROUGE} & \multicolumn{1}{l}{\mustardcell BERTS} & \multicolumn{1}{l}{\peachcell $\mathcal{K}$-BERTS} & \multicolumn{1}{l}{\peachcell $\mathcal{K}$-Precision} \\ \midrule
Flan-T5$_{\textsc{Base}}$ & 21.37 &	34.23 &	60.99	& 70.69 & 69.70 \\
~~~+\approach    & 21.93	& 34.51	& 61.52	& 73.99  &  100.0	                         \\ \cmidrule{1-6}
Flan-T5$_{\textsc{Large}}$    & 23.64	& 37.34 &	62.99 &	72.47 &	81.58	                                             \\
~~~+\approach    & 25.57 &	38.99	& 63.22	& 71.52 &  100.0	                                                 \\ \cmidrule{1-6}
Flan-T5$_{\textsc{XLarge}}$   & 27.94	& 41.30	& 64.83 &	67.01	& 82.35                                                \\
~~~+\approach    & 27.90 &	39.27 &	64.81	& 77.14 & 100.0                                                   \\ \bottomrule
\end{tabular}
}
\caption{Zero-shot results on DoQA \textit{Movies}.}\label{tab: results doqa movies}
\vspace{-1.5mm}
\end{table}

\begin{table}[!ht]
\def\arraystretch{0.9}
\centering
\resizebox{0.99\linewidth}{!}{%
\begin{tabular}{@{}lrrrrr@{}}
\toprule
Model        & \multicolumn{1}{l}{\mustardcell BLEU} & \multicolumn{1}{l}{\mustardcell ROUGE} & \multicolumn{1}{l}{\mustardcell BERTS} & \multicolumn{1}{l}{\peachcell $\mathcal{K}$-BERTS} & \multicolumn{1}{l}{\peachcell $\mathcal{K}$-Precision} \\ \midrule
Flan-T5$_{\textsc{Base}}$ & 23.52 &	34.96 &	62.27	& 64.76 & 100.0	 \\
~~~+\approach    & 22.40	& 32.95	& 61.88	& 79.12  & 100.0	\\ \cmidrule{1-6}
Flan-T5$_{\textsc{Large}}$    & 27.50 &	41.59	& 66.02	& 69.90 &	100.0	                                             \\
~~~+\approach    & 25.27	& 36.10	& 62.29 &	77.35 &  100.0	                                              \\ \cmidrule{1-6}
Flan-T5$_{\textsc{XLarge}}$   & 25.88	& 41.68	& 66.91 & 	66.42  & 100.0	                                              \\
~~~+\approach    & 23.28	& 36.22	& 63.02 &	81.77   & 100.0	                                               \\ \bottomrule
\end{tabular}
}
\caption{Zero-shot results on DoQA \textit{Travel}.}\label{tab: results doqa travel}
\vspace{-1.5mm}
\end{table}

\section{Zero-Shot Results on TopiOCQA}\label{sec: smaller dataset topiocqa}

The results on TopiOCQA when the smaller dataset DoQA is used for \approach fine-tuning are presented in Table~\ref{tab: results topiocqa pretuned doqa}.

\begin{table}[!h]
\def\arraystretch{0.9}
\centering
\resizebox{0.99\linewidth}{!}{%
\begin{tabular}{@{}lrrrrr@{}}
\toprule
Model        & \multicolumn{1}{l}{\mustardcell BLEU} & \multicolumn{1}{l}{\mustardcell ROUGE} & \multicolumn{1}{l}{\mustardcell BERTS} & \multicolumn{1}{l}{\peachcell $\mathcal{K}$-BERTS} & \multicolumn{1}{l}{\peachcell $\mathcal{K}$-Precision} \\ \midrule
Flan-T5$_{\textsc{Base}}$ & 	19.10 &	43.44	& 63.72 &	68.17 &	100.0 \\
~~~+\approach    & 	16.08 &	31.41 &	58.87	& 68.85	& 100.0 \\ \cmidrule{1-6}
Flan-T5$_{\textsc{Large}}$    &  23.26 &	42.0 &	63.64 &	75.83	& 100.0 \\
~~~+\approach    & 	24.47	& 37.16 &	62.31 &		76.33 & 100.0\\ \cmidrule{1-6}
Flan-T5$_{\textsc{XL}}$   &  22.41 &	42.52	& 63.79 &	77.43	& 100.0 \\
~~~+\approach    & 	27.13	& 40.59	& 62.58	& 76.89 &	100.0 \\ \bottomrule
\end{tabular}
}
\caption{Zero-shot results on TopiOCQA when DoQA is used for \approach fine-tuning.}
\label{tab: results topiocqa pretuned doqa}
\vspace{-1.5mm}
\end{table}

\section{Evaluating Faithfulness with \textit{GPT4-Eval}}\label{app:likert}
The 7-point Likert scale used to evaluate faithfulness via GPT4 (i.e., the \textit{GPT4-Eval} evaluation metric) is provided in Table~\ref{tab:likert}.

\begin{table}[!t]
\def\arraystretch{0.9}
\centering
\resizebox{0.999\linewidth}{!}{%
\begin{tabular}{@{}ll@{}}
\toprule
6 & Only information asked for (perfect)                                                                                                                                         \\ \cmidrule{2-2}
5 & \begin{tabular}[c]{@{}l@{}}Information asked for, but also provided additional\\ information that is relevant to the supporting facts (good)\end{tabular}                    \\ \cmidrule{2-2}
4 & \begin{tabular}[c]{@{}l@{}}Follow-up or generic question (No specific information is asked \\ for), agent asked for clarification (not great not terrible)\end{tabular}      \\ \cmidrule{2-2}
3 & \begin{tabular}[c]{@{}l@{}}Information asked for, but also provided additional information \\ that is irrelevant to the supporting facts (not bad)\end{tabular}              \\ \cmidrule{2-2}
2 & Transfer the customer to the correct customer service department (ok)                                                                                                        \\ \cmidrule{2-2}
1 & \begin{tabular}[c]{@{}l@{}}No information asked for, but provided additional information \\ that is either relevant or irrelevant to the supporting facts (bad)\end{tabular} \\ \cmidrule{2-2}
0 & \begin{tabular}[c]{@{}l@{}}Information provided is not coming from the supporting facts \\ (terrible), or transfer customers to the wrong queue (poor)\end{tabular}          \\ \bottomrule
\end{tabular}
}
\caption{Likert-scale for evaluating faithfulness automatically via GPT4.}
\vspace{-1.5mm}
\label{tab:likert}
\end{table}

\section{Results for Faithfulness vs. Abstractiveness}\label{sec:results_faithful_abstractiveness}

\begin{figure}[!t]
    \centering
    \includegraphics[width=0.46\textwidth]{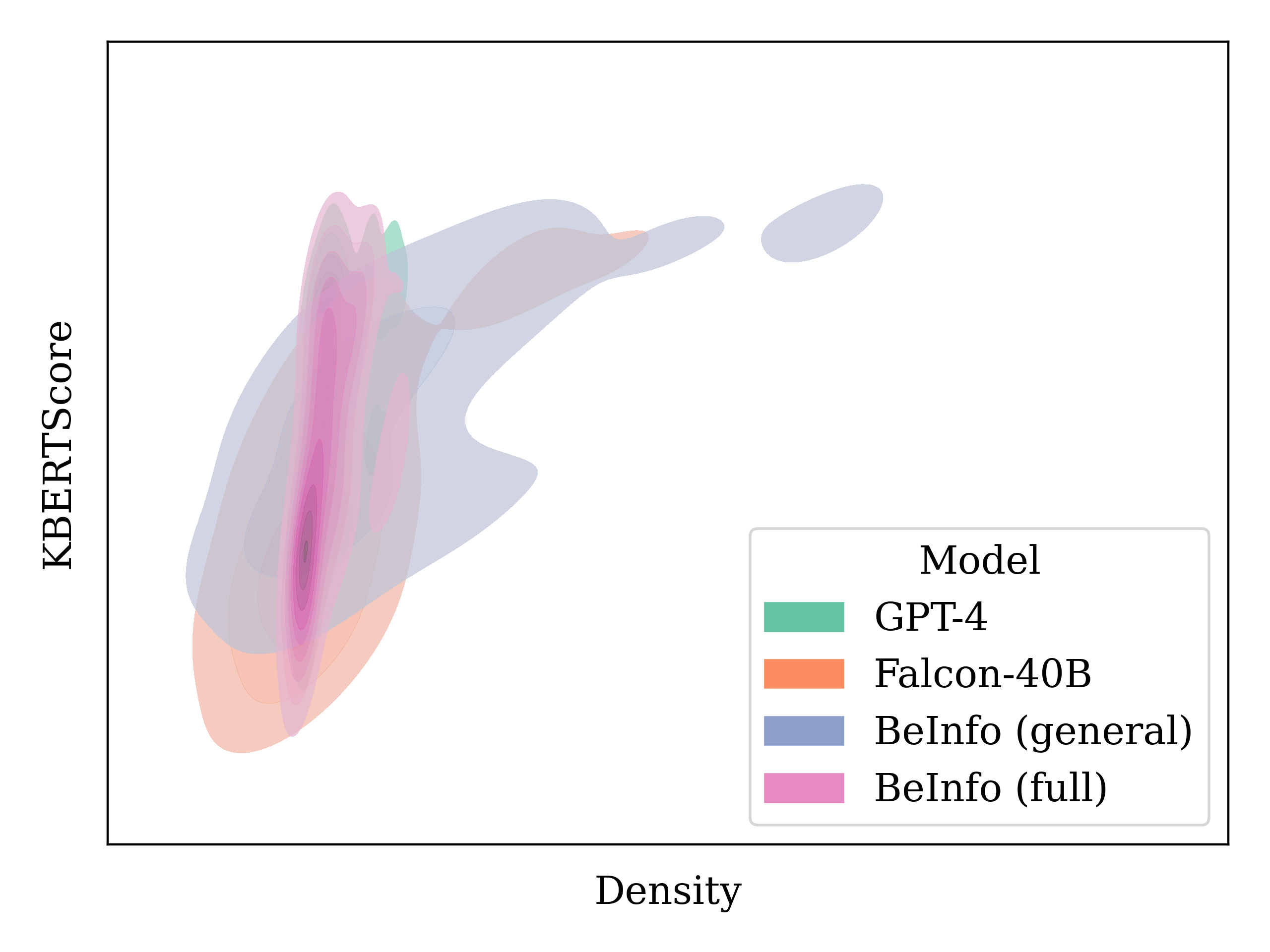}
    \vspace{-0.5mm}
    \caption{Density and $\mathcal{K}$-BERTScore illustrating the trade-off between faithfulness (y-axis) and abstractiveness (x-axis).}
    \label{appfig:faithfulness_vs_abstractiveness}
    \vspace{-1mm}
\end{figure}

The results for factual faithfulness and abstractiveness on real conversations for Flan-T5$_{XL}$ tuned with \approach and larger language models are shown in Figure \ref{appfig:faithfulness_vs_abstractiveness}. Results demonstrate that \approach approximates a much smaller model to the performance of GPT-4 while overcoming the performance of a much larger open-source model, Falcon-40B.  The exact numbers are shown in Table \ref{tab:results_faithful_abstractiveness}.

\begin{table}[!h]
\def\arraystretch{0.9}
\centering
\resizebox{0.99\linewidth}{!}{%
\begin{tabular}{@{}lccc@{}}
\toprule
\rowcolor{Gray}  Model                                                        & Density ($\downarrow$) & Coverage ($\downarrow$) & $\mathcal{K}$-BERTScore ($\uparrow$) \\ \midrule
Flan-T5-XL      &  4.03  &    0.47  &    83.51   \\
\approach (t)     &  2.35  &   49.18   &    84.64   \\
\approach (g)                                                   & 12.10   & 0.73     & 88.33      \\
\approach (f)                                               & 2.32    & 0.60     & 86.30      \\
Falcon-40B                                                   & 5.72    & 0.46     & 84.25      \\
GPT-4                                                        & 2.01    & 0.64     & 87.49      \\ \bottomrule
\end{tabular}
}
\caption{Results for faithfulness and abstractiveness on real user conversations. We use: a) $\mathcal{K}$-BERTScore to measure faithfulness of the model to the knowledge source $\mathcal{K}$; b) Density and Coverage \cite{grusky-etal-2018-newsroom} to measure abstractiveness of the responses. (t)=task-tuned; (g)=general-only; (f)=full. }\label{tab:results_faithful_abstractiveness}
\end{table}

\end{document}